\documentclass[]{article} 
\usepackage{iclr2015,times}
\usepackage{hyperref}
\usepackage{url}
\usepackage{epsfig}
\usepackage{graphicx} 

\title{Object detectors emerge in Deep Scene CNNs}

\author{
Bolei Zhou, Aditya Khosla, Agata Lapedriza, Aude Oliva, Antonio Torralba \\
Computer Science and Artificial Intelligence Laboratory, MIT\\
\texttt{\{bolei,khosla,agata,oliva,torralba\}@mit.edu}
}

\iclrfinalcopy 

\iclrconference 

\begin{document}

\maketitle
\begin{abstract}

With the success of new computational architectures for visual processing, such as convolutional neural networks (CNN) and access to image databases with millions of labeled examples (e.g., ImageNet, Places), the state of the art in computer vision is advancing rapidly. One important factor for continued progress is to understand the representations that are learned by the inner layers of these deep architectures. Here we show that object detectors emerge from training CNNs to perform scene classification. As scenes are composed of objects, the CNN for scene classification automatically discovers meaningful objects detectors, representative of the learned scene categories. With object detectors emerging as a result of learning to recognize scenes, our work demonstrates that the same network can perform both scene recognition and object localization in a single forward-pass, without ever having been explicitly taught the notion of objects.

\end{abstract}

\section{Introduction}

Current deep neural networks achieve remarkable performance at a number of vision tasks surpassing techniques based on hand-crafted features. However, while  the structure of the representation in hand-crafted features is often clear and interpretable, in the case of deep networks it remains unclear what the nature of the learned representation is and why it works so well. A convolutional neural network (CNN) trained on ImageNet~\citep{deng2009imagenet} significantly outperforms the best hand crafted features on the ImageNet challenge ~\citep{ILSVRCarxiv14}. But more surprisingly, the same network, when used as a generic feature extractor, is also very successful at other tasks like object detection on the PASCAL VOC dataset~\citep{Everingham10}. 

A number of works have focused on understanding the representation learned by CNNs. The work by \cite{Zeiler14} introduces a procedure to visualize what activates each unit. Recently \citet{yosinski2014transferable} use transfer learning to measure how generic/specific the learned features are. In \citet{agrawal2014analyzing} and \citet{szegedy2013intriguing}, they suggest that the CNN for ImageNet learns a distributed code for objects. They all use ImageNet, an object-centric dataset, as a training set.

When training a CNN to distinguish different object classes, it is unclear what the underlying representation should be. Objects have often been described using part-based representations where parts can be shared across objects, forming a distributed code. However, what those parts should be is unclear. For instance, one would think that the meaningful parts of a face are the mouth, the two eyes, and the nose. However, those are simply functional parts, with words associated with them; the object parts that are important for visual recognition might be different from these semantic parts, making it difficult to evaluate how efficient a representation is. In fact, the strong internal configuration of objects makes the definition of what is a useful part poorly constrained: an algorithm can find different and arbitrary part configurations, all giving similar recognition performance.

Learning to classify scenes (i.e., classifying an image as being an office, a restaurant, a street, etc) using the Places dataset \citep{zhou2014learning} gives the opportunity to study the internal representation learned by a CNN on a task other than object recognition.


In the case of scenes, the representation is clearer. Scene categories are defined by the objects they contain and, to some extent, by the spatial configuration of those objects. For instance, the important parts of a bedroom are the bed, a side table, a lamp, a cabinet, as well as the walls, floor and ceiling. Objects represent therefore a distributed code for scenes (i.e., object classes are shared across different scene categories). Importantly, in scenes, the spatial configuration of objects, although compact, has a much larger degree of freedom. It is this loose spatial dependency that, we believe, makes scene representation different from most object classes (most object classes do not have a loose interaction between parts). In addition to objects, other feature regularities of scene categories allow for other representations to emerge, such as textures~\citep{walkermalik}, GIST~\citep{oliva2006building}, bag-of-words~\citep{lazebnik2006beyond}, part-based models ~\citep{pandey2011scene}, and 
ObjectBank~\citep{li2010object}. While a CNN has enough flexibility to learn any of those representations, if meaningful objects emerge without supervision inside the inner layers of the CNN, there will be little ambiguity as to which type of representation these networks are learning.

The main contribution of this paper is to show that object detection emerges inside a CNN trained to recognize scenes, even more than when trained with ImageNet. This is surprising because our results demonstrate that reliable object detectors are found even though, unlike ImageNet, no supervision is provided for objects. Although object discovery with deep neural networks has been shown before in an unsupervised setting~\citep{le2013building}, here we find that many more objects can be naturally discovered, in a supervised setting tuned to scene classification rather than object classification. 

Importantly, the emergence of object detectors inside the CNN suggests that a single network can support recognition at several levels of abstraction (e.g., edges, texture, objects, and scenes) without needing multiple outputs or a collection of networks.  Whereas other works have shown that one can detect objects by applying the network multiple times in different locations~\citep{girshick2014rich}, or focusing attention~\citep{ruslanNips2014}, or by doing segmentation~\citep{Grangier09,Farabet13}, here we show that the same network can do both object localization and scene recognition in a single forward-pass. Another set of recent works~\citep{oquab2014weakly, bergamo2014self} demonstrate the ability of deep networks trained on object classification to do localization without bounding box supervision. However, unlike our work, these require object-level supervision while we only use scenes.

\section{ImageNet-CNN and Places-CNN}
\label{sec:cnns}

Convolutional neural networks have recently obtained astonishing performance on object classification~\citep{krizhevsky2012imagenet} and scene classification~\citep{zhou2014learning}. The ImageNet-CNN from~\citet{Jia13caffe} is trained on 1.3 million images from 1000 object categories of ImageNet (ILSVRC 2012) and achieves a top-1 accuracy of $57.4\%$. With the same network architecture, Places-CNN is trained on 2.4 million images from 205 scene categories of Places Database~\citep{zhou2014learning}, and achieves a top-1 accuracy of $50.0\%$. The network architecture used for both CNNs, as proposed in~\citep{krizhevsky2012imagenet}, is summarized in Table~\ref{network}\footnote{We use \textit{unit} to refer to neurons in the various layers and \textit{features} to refer to their activations.}. Both networks are trained from scratch using only the specified dataset.

The deep features from Places-CNN tend to perform better on scene-related recognition tasks compared to the features from ImageNet-CNN. For example, as compared to the Places-CNN that achieves 50.0\% on scene classification, the ImageNet-CNN combined with a linear SVM only achieves $40.8\%$ on the same test set\footnote{Scene recognition demo of Places-CNN is available at \url{http://places.csail.mit.edu/demo.html}. The demo has 77.3\% top-5 recognition rate in the wild estimated from 968 anonymous user responses.} illustrating the importance of having scene-centric data.



\begin{table}\caption{The parameters of the network architecture used for ImageNet-CNN and Places-CNN.}\label{network}
\footnotesize
\centering
\begin{tabular}{ccccccccccc}
\hline
\hline
Layer &conv1 & pool1 & conv2 & pool2& conv3& conv4& conv5& pool5& fc6 & fc7\\
\hline
Units & 96 & 96 & 256 & 256 & 384 & 384 & 256 & 256 & 4096 & 4096\\
Feature & 55$\times$55& 27$\times$27 & 27$\times$27 & 13$\times$13 & 13$\times$13 & 13$\times$13& 13$\times$13 & 6$\times$6 & 1 & 1 \\
\hline
\end{tabular}
\vspace*{-4mm}
\end{table}



\begin{figure}
\begin{center}
\includegraphics[width=1\textwidth]{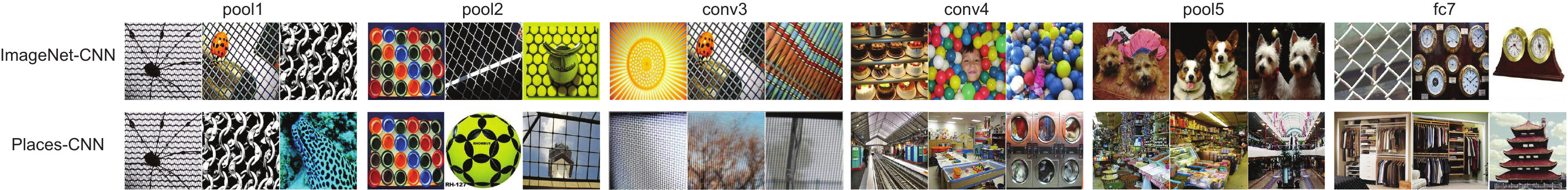}
\end{center}
\caption{Top 3 images producing the largest activation of units in each layer of ImageNet-CNN (top) and Places-CNN (bottom).}
\label{fig:preferred}
\end{figure}

To further highlight the difference in representations, we conduct a simple experiment to identify the differences in the type of images preferred at the different layers of each network: we create a set of 200k images with an approximately equal distribution of scene-centric and object-centric images\footnote{100k object-centric images from the test set of ImageNet LSVRC2012 and 108k scene-centric images from the SUN dataset~\citep{SUNDBijcv}.}, and run them through both networks, recording the activations at each layer. For each layer, we obtain the top 100 images that have the largest average activation (sum over all spatial locations for a given layer). Fig.~\ref{fig:preferred} shows the top 3 images for each layer. We observe that the earlier layers such as pool1 and pool2 prefer similar images for both networks while the later layers tend to be more specialized to the specific task of scene or object categorization. For layer pool2, $55\%$ and $47\%$ of the top-100 images belong to the ImageNet dataset 
for ImageNet-CNN and Places-CNN. Starting from layer conv4, we observe a significant difference in the number of top-100 
belonging to each dataset corresponding to each network. For fc7, we observe that $78\%$ and $24\%$ of the top-100 images belong to the ImageNet dataset for the ImageNet-CNN and Places-CNN respectively, illustrating a clear bias in each network.

In the following sections, we further investigate the differences between these networks, and focus on better understanding the nature of the representation learned by Places-CNN when doing scene classification in order to clarify some part of the secret to their great performance.









\section{Uncovering the CNN representation}

The performance of scene recognition using Places-CNN is quite impressive given the difficulty of the task. In this section, our goal is to understand the nature of the representation that the network is learning.

\subsection{Simplifying the  input images}


Simplifying images is a well known strategy to test human recognition. For example, one can remove information from the image to test if it is diagnostic or not of a particular object or scene (for a review see~\citet{biederman95}). A similar procedure was also used by ~\citet{tanaka93} to understand the receptive fields of complex cells in the inferior temporal cortex (IT).

Inspired by these approaches, our idea is the following: given an image that is correctly classified by the network, we want to simplify this image such that it keeps as little visual information as possible while still having a high classification score for the same category. This simplified image (named minimal image representation) will allow us to highlight the elements that lead to the high classification score. In order to do this, we manipulate images in the gradient space as typically done in computer graphics~\citep{Perez03}. We investigate two different approaches described below.

\begin{figure}
\begin{center}
\includegraphics[width=1\textwidth]{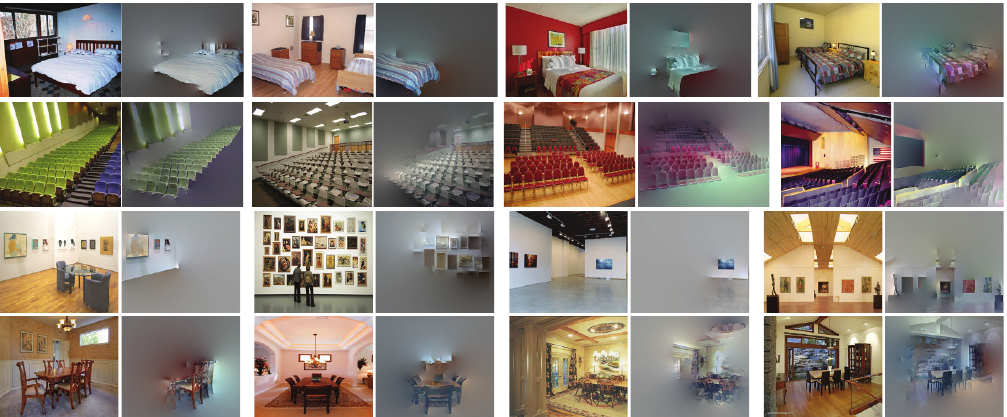}
\end{center}
\vspace{-6mm}
\caption{Each pair of images shows the original image (left) and a simplified image (right) that gets classified by the Places-CNN as the same scene category as the original image. From top to bottom, the four rows show different scene categories: bedroom, auditorium, art gallery, and dining room.}
\label{fig:simplified}
\end{figure}

In the first approach, given an image, we create a segmentation of edges and regions and remove segments from the image iteratively. At each iteration we remove the segment that produces the smallest decrease of the correct classification score and we do this until the image is incorrectly classified. At the end, we get a representation of the original image that contains, approximately, the minimal amount of information needed by the network to correctly recognize the scene category. In Fig.~\ref{fig:simplified} we show some examples of these minimal image representations. Notice that objects seem to contribute important information for the network to recognize the scene. For instance, in the case of bedrooms these minimal image representations usually contain the region of the bed, or in the art gallery category, the regions of the paintings on the walls.

Based on the previous results, we hypothesized that for the Places-CNN, some objects were crucial for recognizing scenes. This inspired our second approach: we generate the minimal image representations using the fully annotated image set of SUN Database~\citep{SUNDBijcv} (see section~\ref {sec_what_object_classes} for details on this dataset) instead of performing automatic segmentation. We follow the same procedure as the first approach using the ground-truth object segments provided in the database.

This led to some interesting observations: for bedrooms, the minimal representations retained the bed in $87\%$ of the cases. Other objects kept in bedrooms were wall ($28\%$) and window ($21\%$). For art gallery the minimal image representations contained paintings ($81\%$) and pictures ($58\%$); in amusement parks, carousel ($75\%$), ride ($64\%$), and roller coaster ($50\%$); in bookstore, bookcase ($96\%$), books ($68\%$), and shelves ($67\%$). These results suggest that object detection is an important part of the representation built by the network to obtain discriminative information for scene classification.



\subsection{Visualizing the receptive fields of units and their activation patterns}
\label{sec:viz}

\begin{figure}
\begin{center}
\includegraphics[width=1\linewidth]{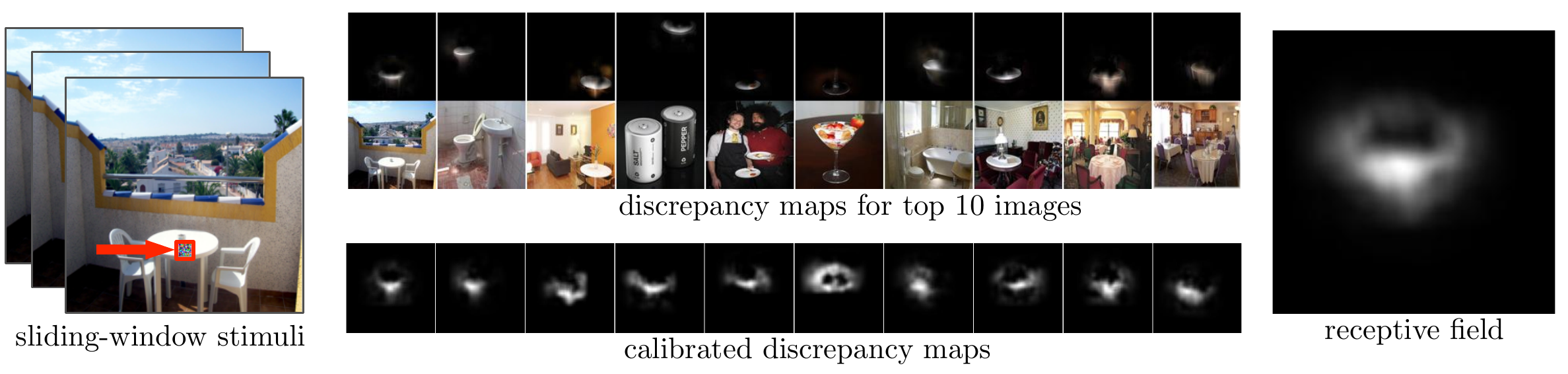}
\end{center}
\vspace{-6mm}
\caption{The pipeline for estimating the RF of each unit. Each sliding-window stimuli contains a small randomized patch (example indicated by red arrow) at different spatial locations. By comparing the activation response of the sliding-window stimuli with the activation response of the original image, we obtain a discrepancy map for each image (middle top). By summing up the calibrated discrepancy maps (middle bottom) for the top ranked images, we obtain the actual RF of that unit (right).}
\label{pipeline_rf}
\end{figure}

In this section, we investigate the shape and size of the receptive fields (RFs) of the various units in the CNNs. While theoretical RF sizes can be computed given the network architecture~\citep{LongNIPS14}, we are interested in the actual, or \textit{empirical} size of the RFs. We expect the empirical RFs to be better localized and more representative of the information they capture than the theoretical ones, allowing us to better understand what is learned by each unit of the CNN.


Thus, we propose a data-driven approach to estimate the learned RF of each unit in each layer. It is simpler than the deconvolutional network visualization method~\citep{Zeiler14} and can be easily extended to visualize any learned CNNs\footnote{More visualizations are available at \url{http://places.csail.mit.edu/visualization}}.


The procedure for estimating a given unit's RF, as illustrated in Fig.~\ref{pipeline_rf}, is as follows. As input, we use an image set of 200k images with a roughly equal distribution of scenes and objects (similar to Sec.~\ref{sec:cnns}). Then, we select the top $K$ images with the highest activations for the given unit. 

For each of the $K$ images, we now want to identify exactly which regions of the image lead to the high unit activations. To do this, we replicate each image many times with small random occluders (image patches of size 11$\times$11) at different locations in the image. Specifically, we generate occluders in a dense grid with a stride of 3. This results in about 5000 occluded images per original image. We now feed all the occluded images into the same network and record the change in activation as compared to using the original image. If there is a large discrepancy, we know that the given patch is important and vice versa. This allows us to build a discrepancy map for each image. 

Finally, to consolidate the information from the $K$ images, we center the discrepancy map around the spatial location of the unit that caused the maximum activation for the given image. Then we average the re-centered discrepancy maps to generate the final RF.

\begin{figure}
\begin{center}
\includegraphics[width=1\linewidth]{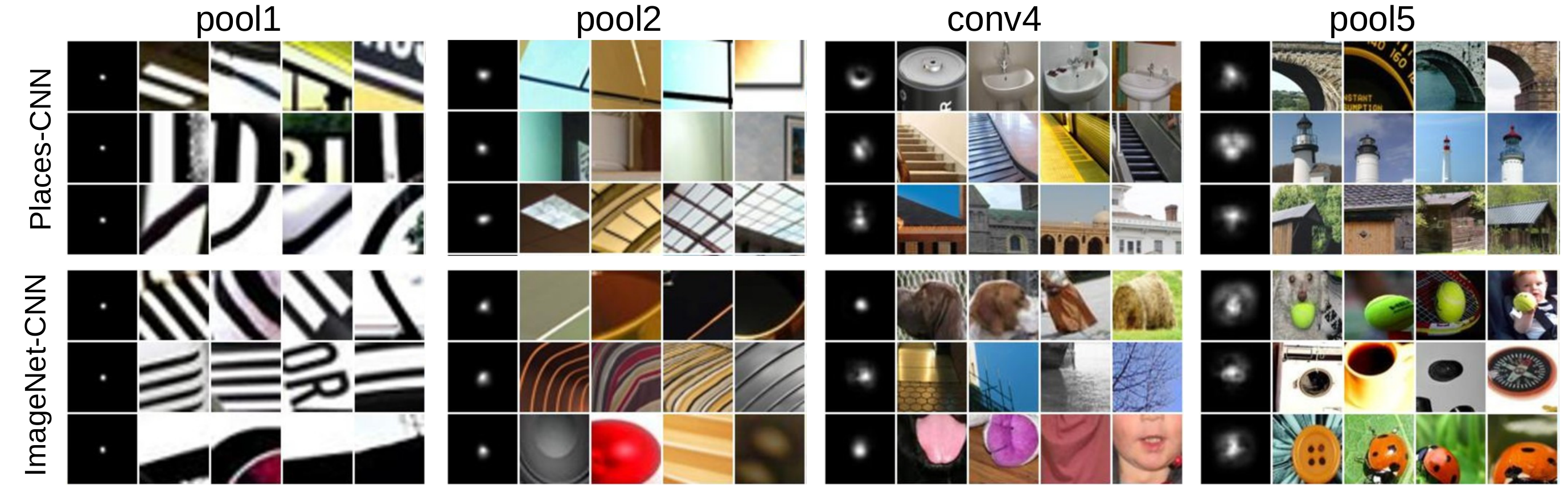}
\end{center}
\caption{The RFs of 3 units of pool1, pool2, conv4, and pool5 layers respectively for ImageNet- and Places-CNNs, along with the image patches corresponding to the top activation regions inside the RFs.}\label{plot_rf}
\end{figure}
\begin{figure}
\begin{center}
\includegraphics[width=1\textwidth]{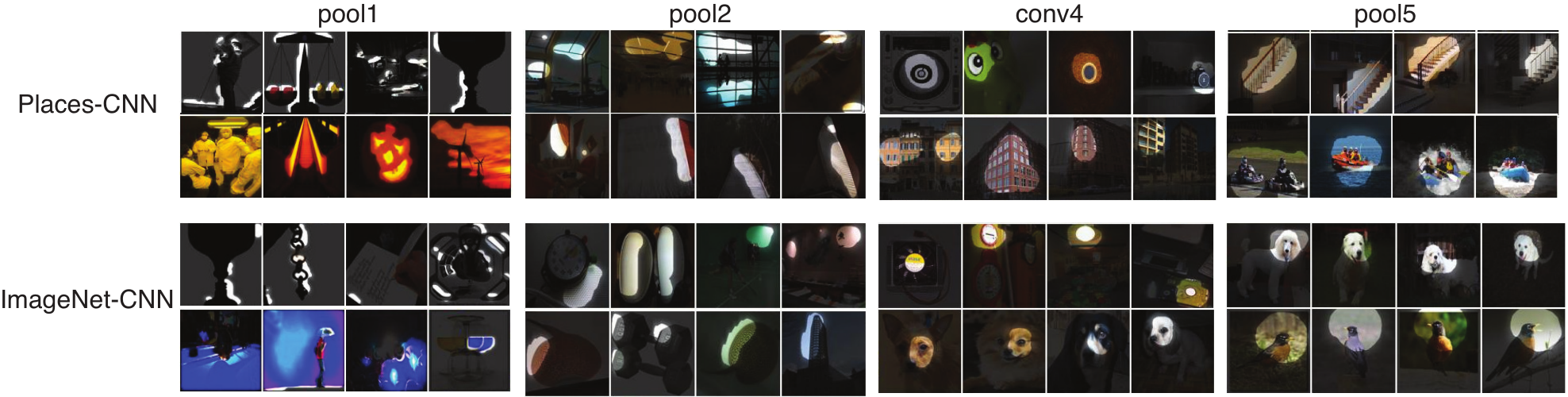}
\end{center}
\vspace{-5mm}
\caption{Segmentation based on RFs. Each row shows the 4 most confident images for some unit.}\label{rf_segmentation}
\end{figure}

In Fig.~\ref{plot_rf} we visualize the RFs for units from 4 different layers of the Places-CNN and ImageNet-CNN, along with their highest scoring activation regions inside the RF. We observe that, as the layers go deeper, the RF size gradually increases and the activation regions become more semantically meaningful. Further, as shown in Fig.~\ref{rf_segmentation}, we use the RFs to segment images using the feature maps of different units. Lastly, in Table~\ref{receptivefield}, we compare the theoretical and empirical size of the RFs at different layers. As expected, the actual size of the RF is much smaller than the theoretical size, especially in the later layers. Overall, this analysis allows us to better understand each unit by focusing precisely on the important regions of each image.

\begin{table}\caption{Comparison of the theoretical and empirical sizes of the RFs for Places-CNN and ImageNet-CNN at different layers. Note that the RFs are assumed to be square shaped, and the sizes reported below are the length of each side of this square, in pixels.}\label{receptivefield}
\centering
\begin{tabular}{lccccc}
\hline
\hline
 &pool1 & pool2 & conv3 & conv4 & pool5 \\
\hline
Theoretic size & 19 & 67 & 99 & 131 & 195\\
Places-CNN actual size & 17.8$\pm$ 1.6 & 37.4$\pm$ 5.9 & 52.1$\pm$10.6 & 60.0$\pm$ 13.7 & 72.0$\pm$ 20.0 \\
ImageNet-CNN actual size & 17.9$\pm$ 1.6 & 36.7$\pm$ 5.4 & 51.1$\pm$9.9 & 60.4$\pm$ 16.0 & 70.3$\pm$ 21.6\\
\hline
\end{tabular}
\end{table}

\begin{figure}
\begin{center}
\includegraphics[width=1\textwidth]{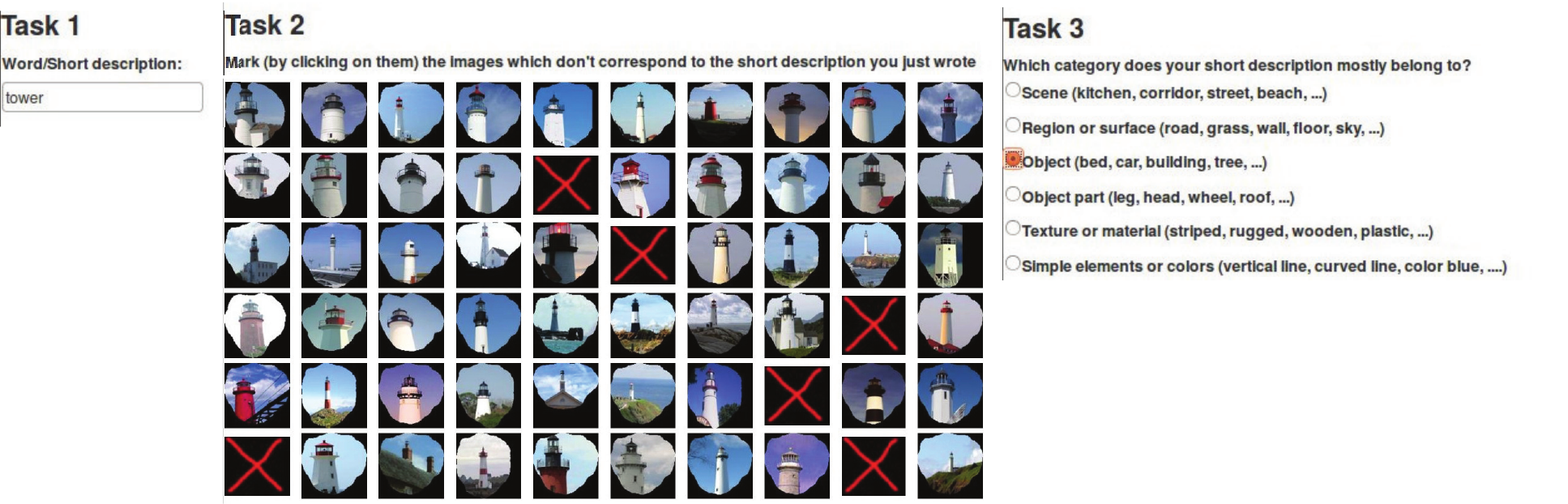}
\end{center}
\vspace{-5mm}
\caption{AMT interface for unit concept annotation. There are three tasks in each annotation.}
\label{fig:AMT}
\end{figure}

\subsection{Identifying the semantics of internal units}

In Section~\ref{sec:viz}, we found the exact RFs of units and observed that activation regions tended to become more semantically meaningful with increasing depth of layers. In this section, our goal is to understand and quantify the precise semantics learned by each unit. 

In order to do this, we ask workers on Amazon Mechanical Turk (AMT) to identify the common theme or \textit{concept} that exists between the top scoring segmentations for each unit. We expect the tags provided by naive annotators to reduce biases. Workers provide tags without being constrained to a dictionary of terms that could bias or limit the identification of interesting properties. 

Specifically, we divide the task into three main steps as shown in Fig.~\ref{fig:AMT}. We show workers the top 60 segmented images that most strongly activate one unit and we ask them to (1) identify the concept, or semantic theme given by the set of 60  images e.g., car, blue, vertical lines, etc, (2) mark the set of images that do not fall into this theme, and (3) categorize the concept provided in (1) to one of 6 semantic groups ranging from low-level to high-level: simple elements and colors (e.g., horizontal lines, blue), materials and textures (e.g., wood, square grid), regions ans surfaces (e.g., road, grass), object parts (e.g., head, leg), objects (e.g., car, person), and scenes (e.g., kitchen, corridor). This allows us to obtain both the semantic information for each unit, as well as the level of abstraction provided by the labeled concept. 

\begin{figure}
\begin{center}
\includegraphics[width=1\textwidth]{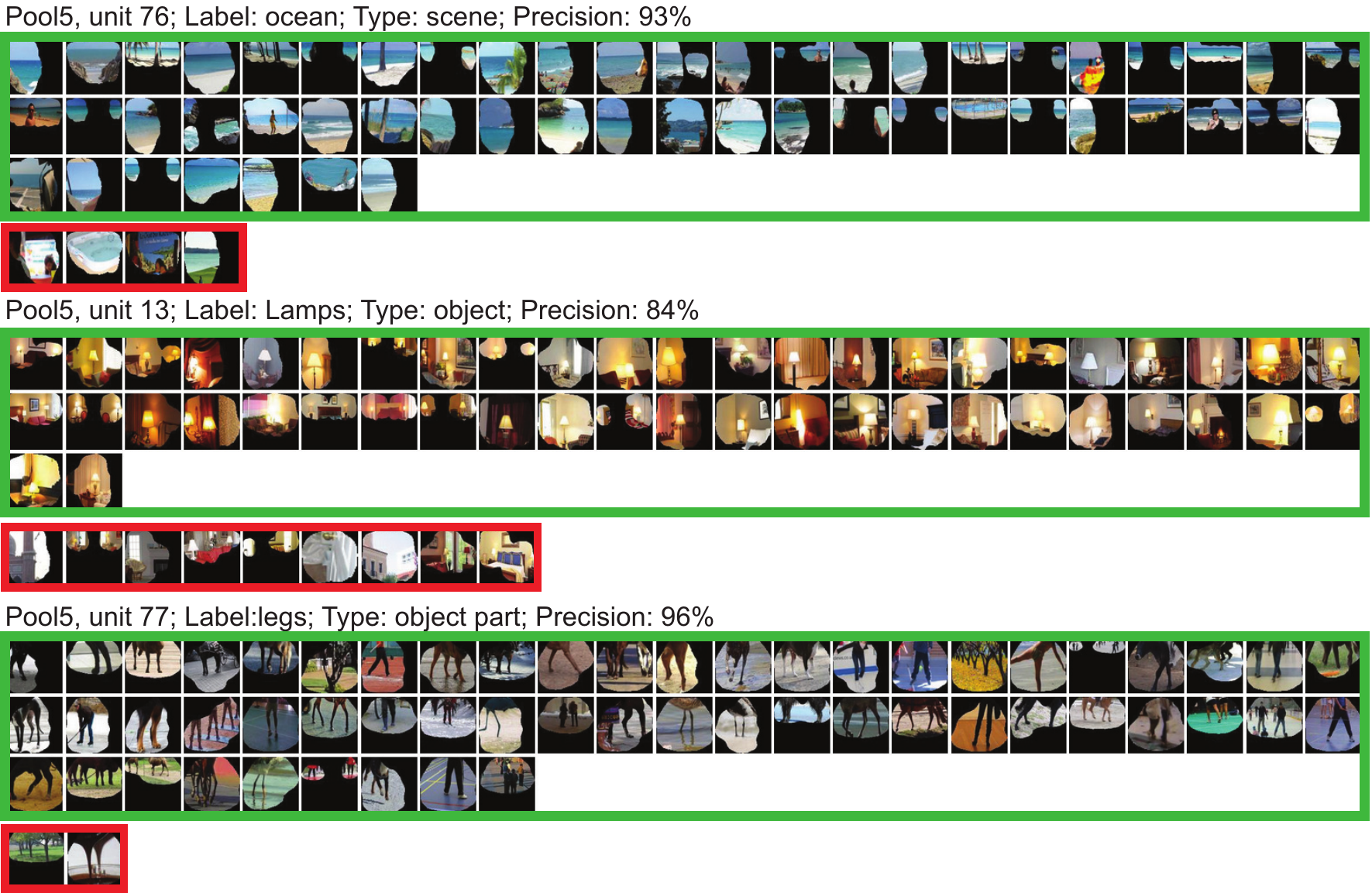}
\includegraphics[width=1\textwidth]{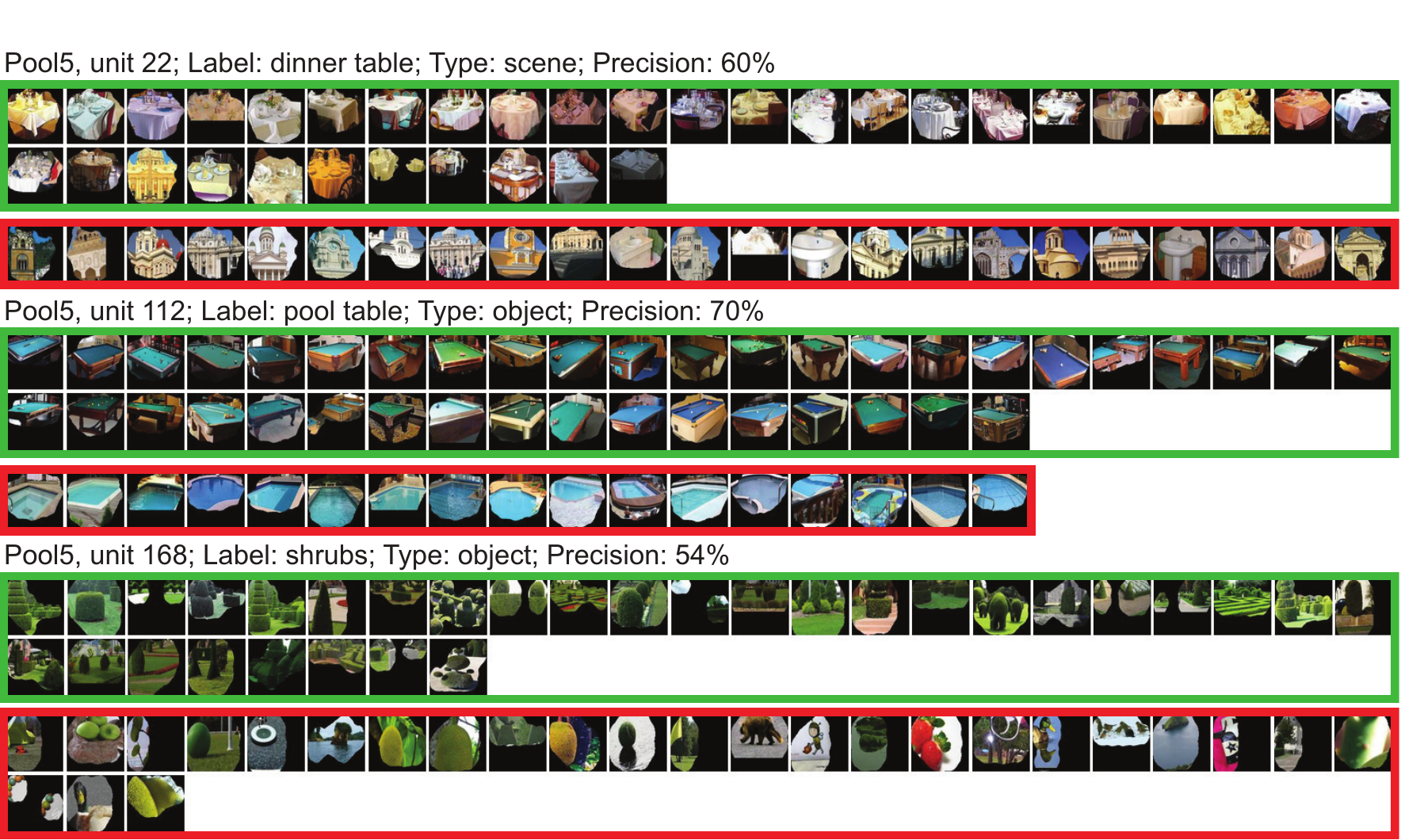}
\end{center}
\caption{Examples of unit annotations provided by AMT workers for 6 units from pool5 in Places-CNN. For each unit the figure shows the label provided by the worker, the type of label, the images selected as corresponding to the concept (green box) and the images marked as incorrect (red box). The precision is the percentage of correct images. The top three units have high performance while the bottom three have low performance ($<75\%$).}
\label{fig:semantics}
\end{figure}

\begin{figure}[t]
\begin{center}
\includegraphics[width=1\textwidth]{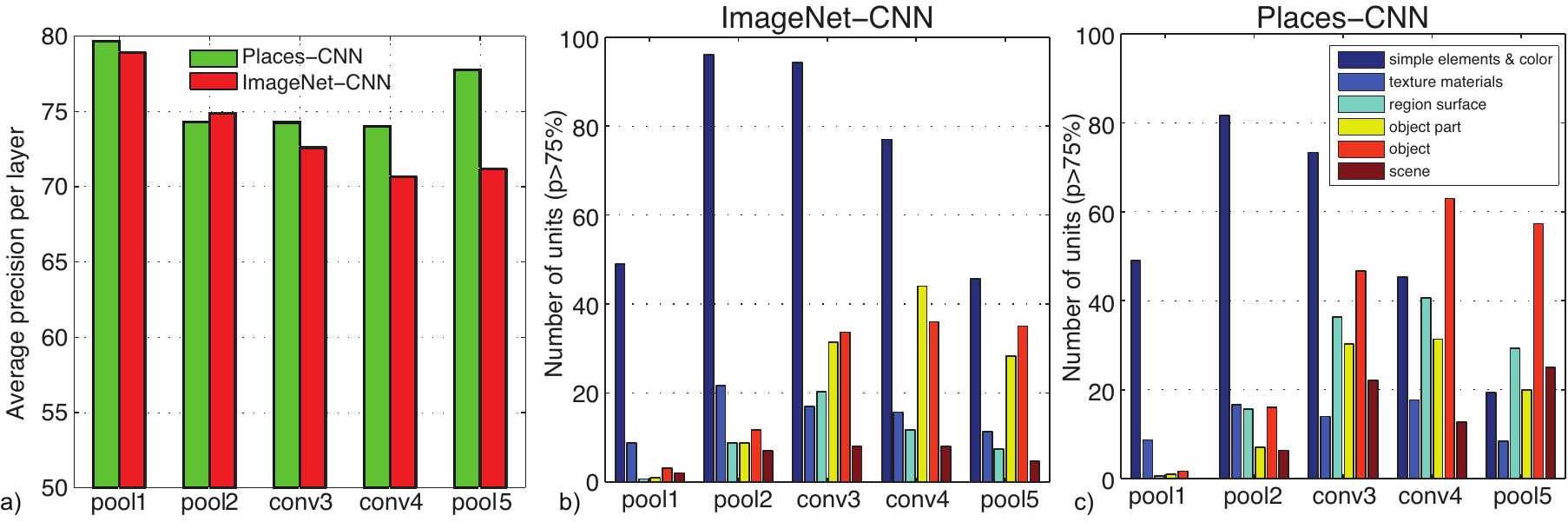}
\end{center}
\vspace*{-4mm}
\caption{(a) Average precision of all the units in each layer for both networks as reported by AMT workers. (b) and (c) show the number of units providing different levels of semantics for ImageNet-CNN and Places-CNN respectively.}
\label{fig:distributionsemantics}
\end{figure}

To ensure high quality of annotation, we included 3 images with high negative scores that the workers were required to identify as negatives in order to submit the task. Fig.~\ref{fig:semantics} shows some example annotations by workers. For each unit, we measure its precision as the percentage of images that were selected as fitting the labeled concept. In Fig.~\ref{fig:distributionsemantics}.(a) we plot the average precision for ImageNet-CNN and Places-CNN for each layer.

In Fig.~\ref{fig:distributionsemantics}.(b-c) we plot the distribution of  concept categories for ImageNet-CNN and Places-CNN at each layer. For this plot we consider only units that had a precision above $75\%$ as provided by the AMT workers. Around $60\%$ of the units on each layer where above that threshold. For both networks, units at the early layers (pool1, pool2) have more units responsive to simple elements and colors, while those at later layers (conv4, pool5) have more high-level semantics (responsive more to objects and scenes). Furthermore, we observe that conv4 and pool5 units in Places-CNN have higher ratios of high-level semantics as compared to the units in ImageNet-CNN. 

\begin{figure}[t]
\begin{center}
\includegraphics[width=1\textwidth]{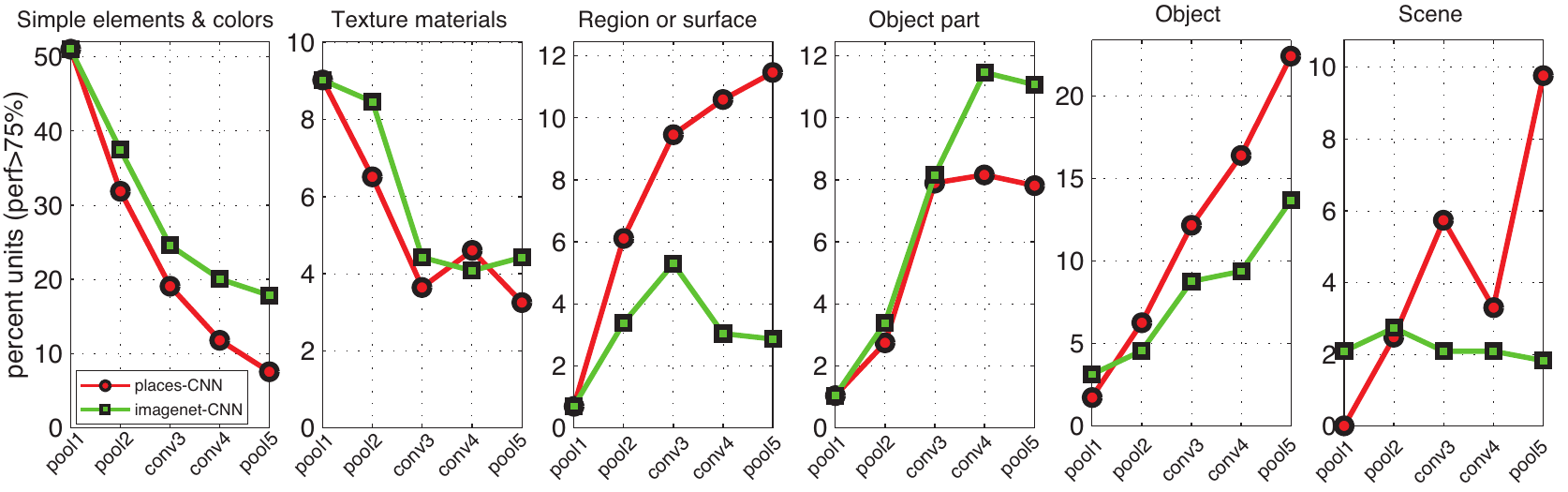}
\end{center}
\vspace*{-4mm}
\caption{Distribution of semantic types found for all the units in both networks. From left to right, each plot corresponds to the distribution of units in each layer assigned to simple elements or colors, textures or materials, regions or surfaces, object parts, objects, and scenes. The vertical axis is the percentage of units with each layer assigned to each type of concept.} 
\label{fig:distributionsemanticslayers}
\end{figure}

Fig.~\ref{fig:distributionsemanticslayers} provides a different visualization of the same data as in Fig.~\ref{fig:distributionsemantics}.(b-c). This plot better reveals how different levels of abstraction emerge in different layers of both networks. The vertical axis indicates the percentage of units in each layer assigned to each concept category. ImageNet-CNN has more units tuned to simple elements and colors than Places-CNN while Places-CNN has more objects and scenes. ImageNet-CNN has more units tuned to object parts (with the maximum around conv4). It is interesting to note that Places-CNN discovers more objects than ImageNet-CNN despite having no object-level supervision.

\section{Emergence of objects as the internal representation}

As shown before, a large number of units in pool5 are devoted to detecting objects and scene-regions (Fig.~\ref{fig:distributionsemanticslayers}). But what categories are found? Is each category mapped to a single unit or are there multiple units for each object class? Can we actually use this information to segment a scene?

\subsection{What object classes emerge?}
\label{sec_what_object_classes}

To answer the question of why certain objects emerge from pool5, we tested ImageNet-CNN and Places-CNN on fully annotated images from the SUN database \citep{SUNDBijcv}. The SUN database contains 8220 fully annotated images from the same 205 place categories used to train Places-CNN. There are no duplicate images between SUN and Places. We use SUN instead of COCO~\citep{mscoco} as we need dense object annotations to study what the most informative object classes for scene categorization are, and what the natural object frequencies in scene images are. For this study, we manually mapped the tags given by AMT workers to the SUN categories. 

\begin{figure}[t]
\begin{center}
\includegraphics[width=1\textwidth]{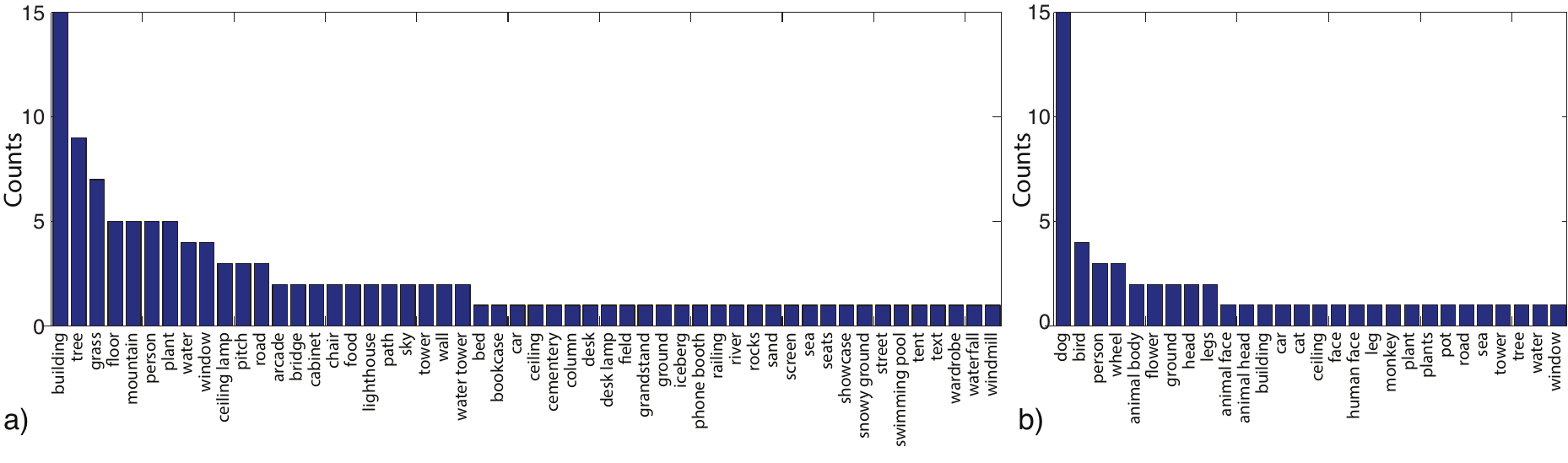}
\end{center}
\vspace*{-8mm}
\caption{Object counts of CNN units discovering each object class for (a) Places-CNN and (b) ImageNet-CNN.} 
\label{fig:comparison}
\end{figure}

\begin{figure}[t]
\begin{center}
\includegraphics[width=1\textwidth]{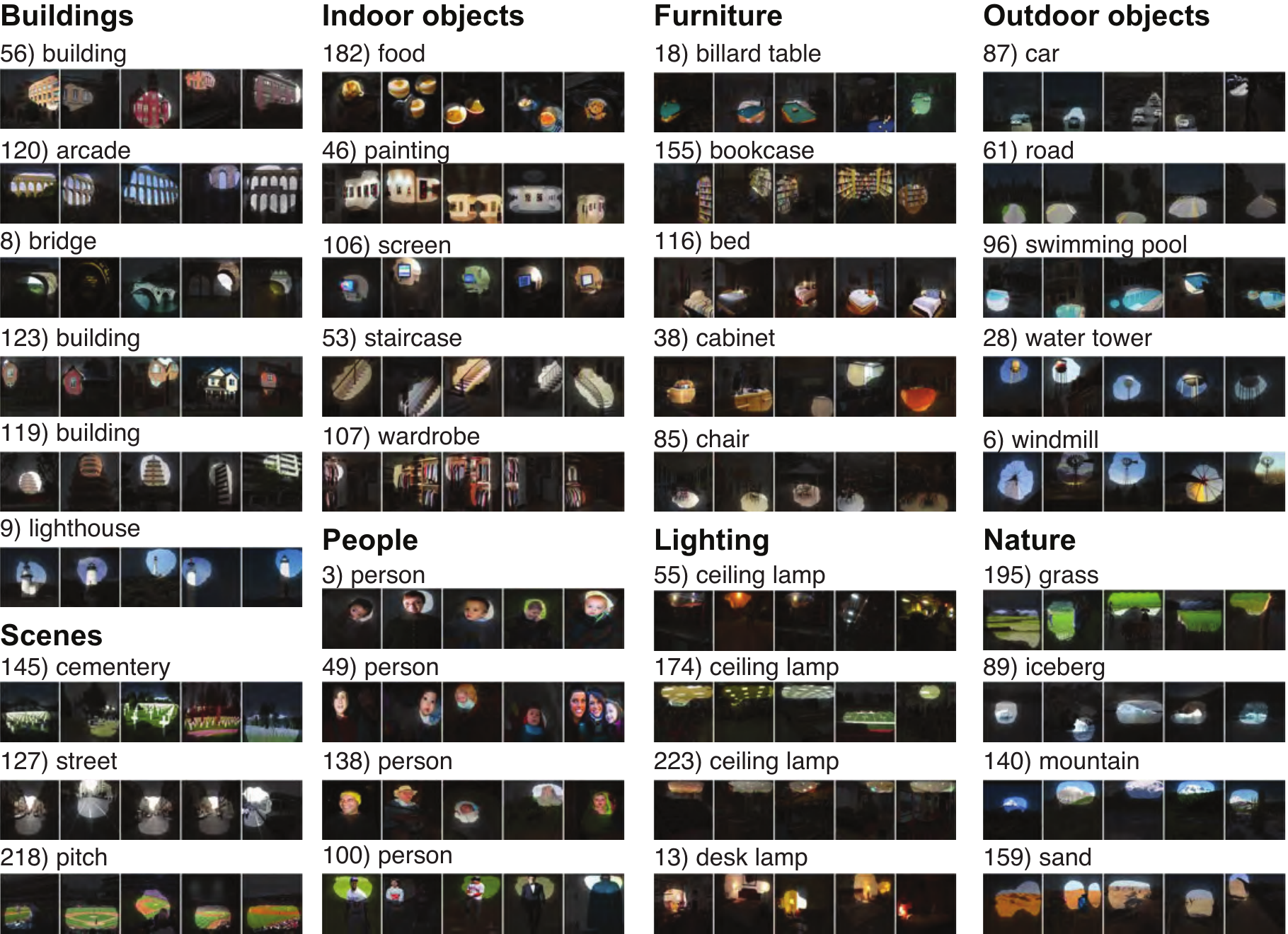}
\end{center}
\vspace*{-6mm}
\caption{Segmentations using pool5 units from Places-CNN. Many classes are encoded by several units covering different object appearances. Each row shows the 5 most confident images for each unit. The number represents the unit number in pool5.}
\label{fig:segmentationsSUN}
\end{figure}

Fig.~\ref{fig:comparison}(a) shows the distribution of objects found in pool5 of Places-CNN. Some objects are detected by several units. For instance, there are 15 units that detect buildings. Fig.~\ref{fig:segmentationsSUN} shows some units from the Places-CNN grouped by the type of object class they seem to be detecting. Each row shows the top five images for a particular unit that produce the strongest activations. The segmentation shows the regions of the image for which the unit is above a certain threshold. Each unit seems to be selective to a particular appearance of the object. For instance, there are 6 units that detect lamps, each unit detecting a particular type of lamp providing finer-grained discrimination; there are 9 units selective to people, each one tuned to different scales or people doing different tasks. 

Fig.~\ref{fig:comparison}(b) shows the distribition of objects found in pool5 of ImageNet-CNN. ImageNet has an abundance of animals among the categories present: in the ImageNet-CNN, out of the 256 units in pool5, there are 15 units devoted to detecting dogs and several more detecting parts of dogs (body, legs, ...). The categories found in pool5 tend to follow the target categories in ImageNet. 

\begin{figure}[t]
\begin{center}
\includegraphics[width=1\textwidth]{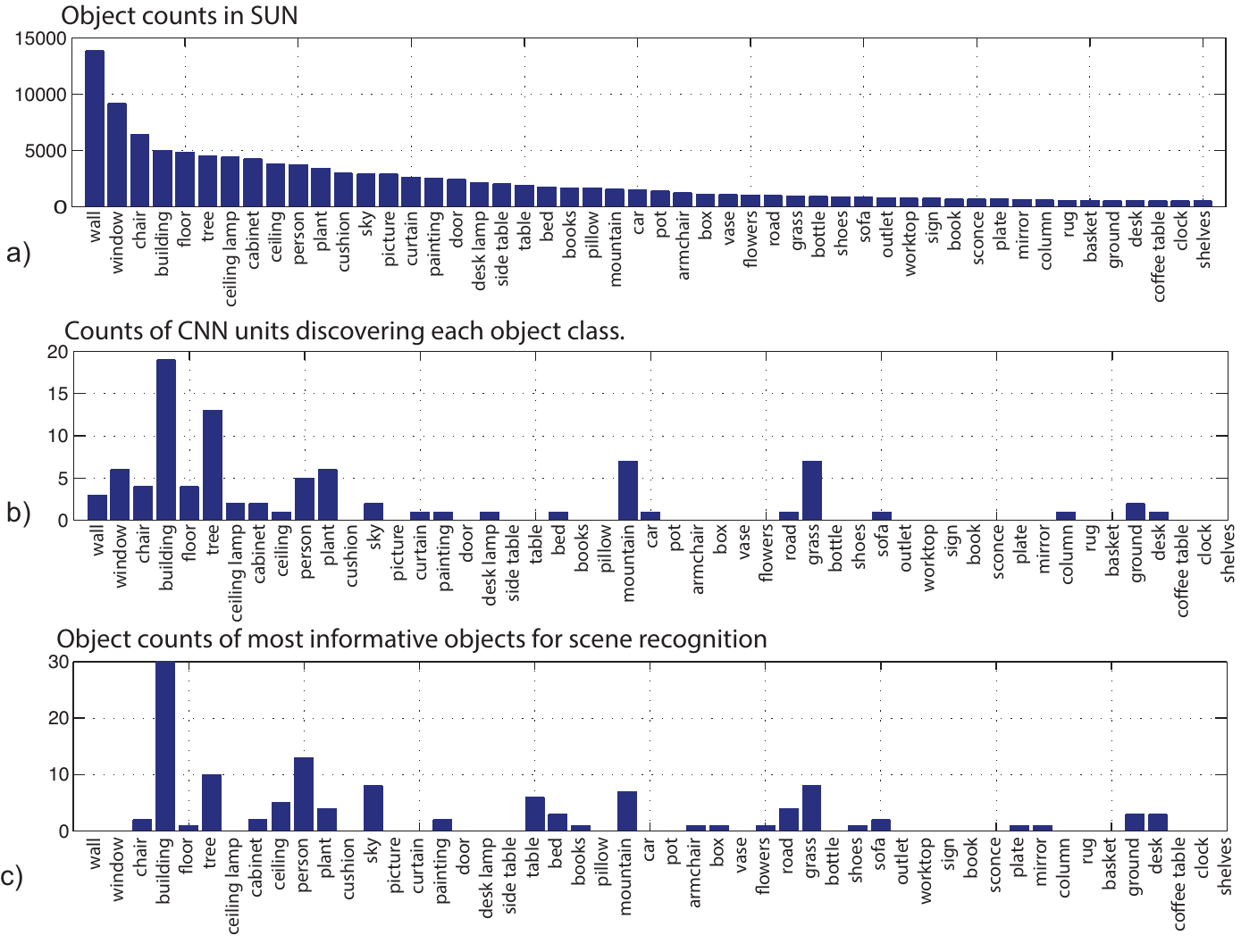}
\end{center}
\vspace{-6mm}
\caption{(a) Object frequency in SUN (only top 50 objects shown), (b) Counts of objects discovered by pool5 in Places-CNN. (c) Frequency of most informative objects for scene classification.}
\label{fig:taxonomy}
\end{figure}

Why do those objects emerge? One possibility is that the objects that emerge in pool5 correspond to the most frequent ones in the database. Fig.~\ref{fig:taxonomy}(a) shows the sorted distribution of object counts in the SUN database which follows Zipf's law. Fig.~\ref{fig:taxonomy}(b) shows the counts of units found in pool5 for each object class (same sorting as in Fig.~\ref{fig:taxonomy}(a)). The correlation between object frequency in the database and object frequency discovered by the units in pool5 is 0.54. Another possibility is that the objects that emerge are the objects that allow discriminating among scene categories. To measure the set of discriminant objects we used the ground truth in the SUN database to measure the classification performance achieved by each object class for scene classification. Then we count how many times each object class appears as the most informative one. This measures the number of scene categories a particular object class is the most useful for. The counts are shown 
in Fig.~\ref{fig:taxonomy}(c). Note the similarity between Fig.~\ref{fig:taxonomy}(b) and Fig.~\ref{fig:taxonomy}(c). The correlation is 0.84 indicating that the network is automatically identifying the most discriminative object categories to a large extent.

Note that there are 115 units in pool5 of Places-CNN not detecting objects. This could be due to incomplete learning or a complementary texture-based or part-based representation of the scenes. Therefore, although objects seem to be a key part of the representation learned by the network, we cannot rule out other representations being used in combination with objects. 

\subsection{Object Localization within the inner Layers}

\begin{figure}[t]
\begin{center}
\includegraphics[width=1\textwidth]{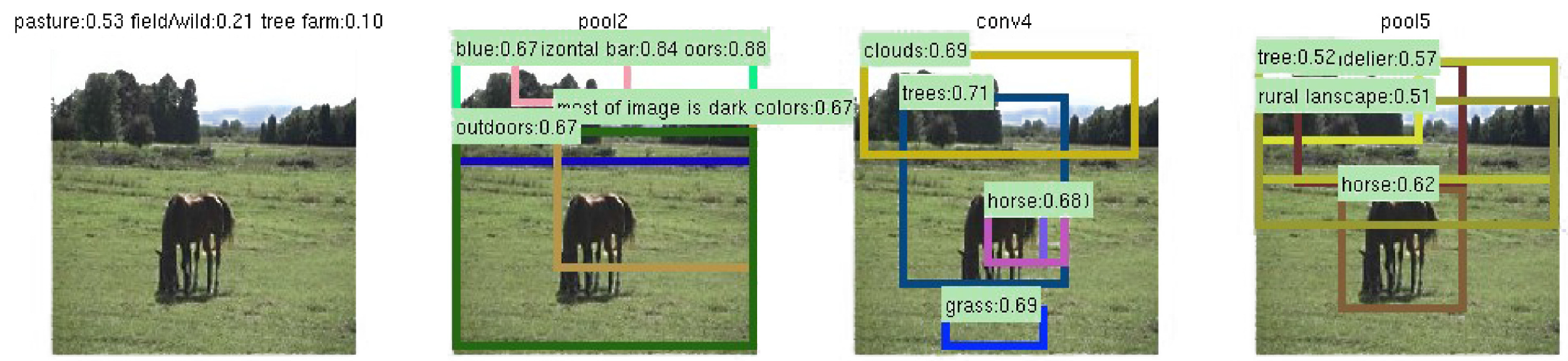}
\end{center}
\vspace{-6mm}
\caption{Interpretation of a picture by different layers of the Places-CNN using the tags provided by AMT workers. The first shows the final layer output of Places-CNN. The other three show detection results along with the confidence based on the units' activation and the semantic tags.}
\label{fig:segLayers}
\end{figure}

Places-CNN is trained to do scene classification using the output of the final layer of logistic regression and achieves state-of-the-art performance. From our analysis above, many of the units in the inner layers could perform interpretable object localization. Thus we could use this single Places-CNN with the annotation of units to do both scene recognition and object localization in a single forward-pass. Fig.~\ref{fig:segLayers} shows an example of the output of different layers of the Places-CNN using the tags provided by AMT workers. Bounding boxes are shown around the areas where each unit is activated within its RF above a certain threshold.

\begin{figure}
\begin{center}
\includegraphics[width=1\textwidth]{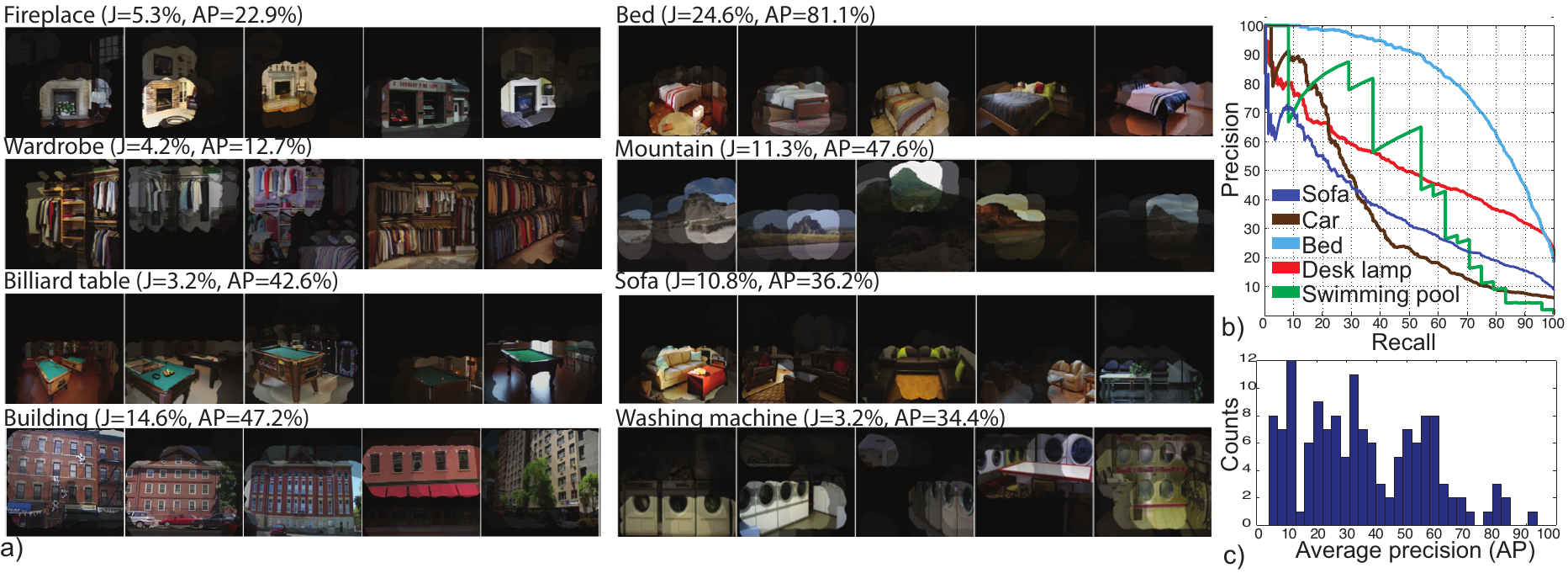}
\end{center}
\vspace{-6mm}
\caption{(a) Segmentation of images from the SUN database using pool5 of Places-CNN (J = Jaccard segmentation index, AP = average precision-recall.) (b) Precision-recall curves for some discovered objects. (c) Histogram of AP for all discovered object classes.}
\label{fig:segmentationSUN}
\end{figure}

In Fig.~\ref{fig:segmentationSUN} we provide the segmentation performance of the objects discovered in pool5 using the SUN database. The performance of many units is very high which provides strong evidence that they are indeed detecting those object classes despite being trained for scene classification.

\section{Conclusion}


We find that object detectors emerge as a result of learning to classify scene categories, showing that a single network can support recognition at several levels of abstraction (e.g., edges, textures, objects, and scenes) without needing multiple outputs or networks. While it is common to train a network to do several tasks and to use the final layer as the output, here we show that reliable outputs can be extracted at each layer. As objects are the parts that compose a scene, detectors tuned to the objects that are discriminant between scenes are learned in the inner layers of the network. Note that only informative objects for specific scene recognition tasks will emerge. Future work should explore which other tasks would allow for other object classes to be learned without the explicit supervision of object labels.

\subsubsection*{Acknowledgments}

This work is supported by the National Science Foundation under Grant No. 1016862 to A.O, ONR MURI N000141010933 to A.T, as well as MIT Big Data Initiative at CSAIL, Google and Xerox Awards, a hardware donation from NVIDIA Corporation, to A.O and A.T.

\bibliography{egbib}
\bibliographystyle{iclr2015}

\end{document}